\documentclass[10pt,twocolumn,letterpaper]{article}

\usepackage{cvpr}
\usepackage{times}
\usepackage{epsfig}
\usepackage{graphicx}
\usepackage{amsmath}
\usepackage{amssymb}

\usepackage[pagebackref=true,breaklinks=true,letterpaper=true,colorlinks,bookmarks=false]{hyperref}

\cvprfinalcopy 

\def\cvprPaperID{1} 

\ifcvprfinal\fi
\begin{document}

\title{Towards an "In-the-Wild" Emotion Dataset Using a Game-based Framework}


\author{
Wei Li\\
Dept of Electrical Engineering\\
CUNY City College\\
New York, USA\\
{\tt\small wli3@ccny.cuny.edu}
\and
Farnaz Abtahi\\
Dept of Computer Science\\
CUNY Graduate Center\\
New York, USA\\
{\tt\small fabtahi@gradcenter.cuny.edu}
\and
Christina Tsangouri\\
Dept of Computer Science\\
CUNY City College\\
New York, USA\\
{\tt\small christinatsangouri@gmail.com}
\and
Zhigang Zhu\\
Dept of Computer Science\\
CUNY Graduate Center and City College\\
New York, USA\\
{\tt\small zhu@cs.ccny.cuny.edu}
}

\maketitle
\begin{abstract}
 In order to create an "in-the-wild" dataset of facial emotions with large number of balanced samples, this paper proposes a game-based data collection framework. The framework mainly include three components: a game engine, a  game interface, and a data collection and evaluation module. We use a deep learning approach to build an emotion classifier as the game engine. Then a emotion web game to allow gamers to enjoy the games, while the data collection module obtains automatically-labelled emotion images. Using our game, we have collected more than 15,000 images within a month of the test run and built an emotion dataset "GaMo". To evaluate the dataset, we compared the performance of two deep learning models trained on both GaMo and CIFE. The results of our experiments show that because of being large and balanced, GaMo can be used to build a more robust emotion detector than the emotion detector trained on CIFE, which was used in the game engine to collect the face images.
\end{abstract}



\section{Introduction}
\textcolor{black}{
Detecting people's emotion has been an interesting research topic for more than 20 years. To analyze people's emotions, high quality datasets are a necessity. Many works have been done to construct useful datasets in the past 20 years. But most of the datasets are either constrained or are not balanced. In this paper, we present a novel "in-the-wild" emotion dataset, the GaMo, using a game-based approach. Comparing to the existing emotion datasets, the new approach can collect a larger number of balanced and "in-the-wild" data, which is more useful in training practical emotion models.} \textcolor{black}{ The dataset is "in-the-wild" since the cameras used, viewing angles, distances, illuminations and ways to make expressions are not constrained when users play the games in their own environments.}

\textcolor{black}{
Among the many datasets that have been provided by researchers for recognizing emotions from images, there are mainly 2 kinds of databases. The first kind are captured in laboratory such as CK+, MMI and DISFA challenge dataset ~\cite{mavadati2013disfa, kanade2000comprehensive, cohn2007observer,lucey2010extended}. Subjects are invited to labs and sitting in lighting and position constrained environment. Good results can be achieved on these datasets but in real life scenario, it's always hard to have a good performance.
The other kind of dataset are collected from existing medias. Such as CIFE and EmotiW ~\cite{li2015deep,li2015emotiw,pantic2005web}. CIFE is a dataset generated by collecting images from web search engines like Google, Baidu and Flickr by using key words corresponding to emotion categories. Using web search engines,  one can easily obtain thousands of images but the dataset are not balanced. Emotiw is a video clip dataset for emotion challenge, and the video samples are from Hollywood movies where the actors show different emotions. For the datasets collected from existing media, parts of the emotions like Happy or Sad are easier to obtain, but for some emotions like Disgust or Fear, it's hard to obtain enough samples. 
}

Although the existing datasets are generally not balanced, many interesting and promising approaches have been proposed for emotion detection. Most existing facial emotion recognition methods have focused on recognizing emotions of frontal faces, such as the images in CK+ \cite{lucey2010extended}. Shan, et al \cite{shan2009facial} have proposed a LBP-based feature extractor combined with an SVM for classification. In the method proposed by Xiao, et al \cite{xiao2011facial}, instead of training one model for all emotions, separate models have been trained for each emotion, which improve the overall performance. Wang, et al \cite{wang2013capturing} modeled facial emotions as complex activities that consist of temporally overlapping sequences of face events. Then, an Interval Temporal Bayesian Network (ITBN) was used to capture the complex temporal information.  Karan, et al ~\cite{sikkaexemplar} proposed a HMM-based approach to make use of consecutive frame information to achieve better emotion recognition accuracy from video.

During the recent years, the multimedia and vision communities have been exposed to the wide spread of Deep Learning methods \cite{taigman2014deepface, sun2014deep}. Deep learning approaches are also used in many emotion detection applications. Liu, et al \cite{liu2014facial} proposed a Boosted Deep Belief Network to perform feature learning, feature selection and classifier construction for emotion recognition. Different DBN models for unsupervised feature learning in audio-visual emotion recognition have been compared in the work done by Kim, et al \cite{kim2013deep}. Li, et al \cite{li2015deep} used CNNs on images collected from the web. To prove the effectiveness of CNNs, they compared their performance on CK+ to the state of the art methods. Multimodal deep learning approaches have been applied to facial emotion recognitions tasks. An example is Jung, et al's work ~\cite{jung2015joint} in which facial landmarks based shape information and image based appearance information are learned through a combined CNN network. The results show that deep learning based multimodal features act better than individual modalities or the use of traditional learning approaches. 

\textcolor{black}{
Deep learning based approaches require a large number of data to train their models. Most of the existing datasets either have a small number of data, or the data samples for different emotion categories are imbalanced. This made it hard for deep learning to reach a robust emotion detector.} A solution to this problem could be to generate a balanced dataset with enough samples of every emotion. Since almost all existing datasets are highly imbalanced, we decided to develop a framework through which we are able to build a new large and balanced dataset. Our approach uses an easy-to-use game based framework, allowing people match on-the-screen faces with different emotions, then their images with various, balanced facial expressions are captured.

In the generation of ImageNet dataset ~\cite{deng2009imagenet}, Amazon Mechanical Turk (AMT) is used to label all the training images. Workers are hired online and can remotely work on labeling the dataset. The ImageNet is a large scale dataset that aims to label 50 million images for object classification and without the help of online workers, the labeling would not be feasible. This inspired us to develop the idea of involving people in data collection process through an online framework, preferable using games. \textcolor{black}{There have been some efforts in using games to attracting people to perform some image classification work. Luis, et al~\cite{von2004labeling} designed an interactive system that attracted people to label images, Mourao et al ~\cite{mourao2013competitive} developed a facial engaging algorithm as the controller to play their NovoEmotions game, and a player engagement dataset was obtained and the relationship between the players' facial engagement and game scores were analyzed. Emotion games have also also used to entertain ASD children and to help them perform facial expressions by mirroring their emotions to some cartoon characters ~\cite{deriso2012emotion}.} Our online framework not only makes use of the online crowd source through games, but also has much lower cost than AMT. And since the numbers of various facial expressions can be controlled by the designs of the games, the dataset can be guaranteed to be balanced. 

The games are simple and straightforward and they work as follows: Anyone can access a game's webpage to play remotely over a web browser. The only requirement is that a camera (either a built-in cam or a webcam) is available. While playing the game, the user will need to make a facial expression to match the emotion that is displayed on the screen at the time. As the emotion is recognized is a match, images are captured and stored in our emotion dataset. Since the high probability that the user makes a facial expression matches the displayed one and is recognized/verified by our emotion recognition program, the manual correction will be minimal (in our data collection, we did not do any cleanup of the data collected by the game). 

\begin{figure}
\centering
\psfig{file=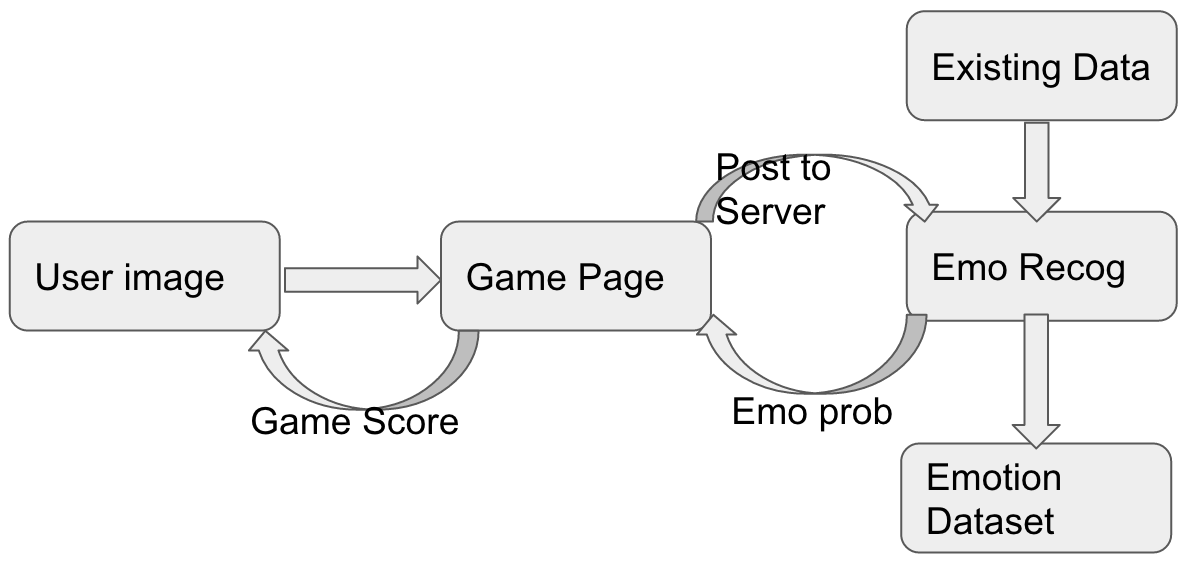, height=1.8in, width=3.0in,}
\caption{Game-based framework for emotion dataset collection.}
\vskip -6pt
\label{fg1}
\end{figure}

Figure \ref{fg1} shows our proposed framework. We first use the CIFE dataset and a  fine-tuned CNN model to build an emotion detector. To decide if the user is making the right face, the user's image is sent to the server and analyzed by our emotion detector and meanwhile saved in our emotion dataset. Since the target emotions are defined by us, we can make the dataset more balanced. The game can be accessed by anyone willing to play, hence we can collect a large scale dataset. The interface has video, images, audio and text and therefor makes it into a fun multimedia game for serious data collection.

Our work has the following main contributions:
1) A low cost multimedia game based data collection framework is proposed; in theory no manual cleanup is needed.
2) We collected a more balanced dataset by designing the engine of the game.
3) The quality of the emotion labels is high since imitation and recognition/verification process.
4) With a dataset consisting of samples that are close to real-life emotions, we can obtain a more accurate model for actual unexaggerated emotions.

 After this section (Section 1) explaining our motivation based some related work and introduces the basic components of our proposed framework, the rest of this paper is organized as follows. Section 2 will describe our game engine, which uses a CNN-based deep learning approach to construct the emotion detector.  Section 3 explains the design of the game interface and the process to obtain a balanced dataset. In Section 4, we evaluate the dataset by comparing it with other datasets through the training and testing of an emotion detector. Finally,  discussions and conclusions  of our work are presented in Section 5.

\section{Game Engine: The Emotion Detector}
The core part of our framework is the model of an emotion detector, since we need it to be a first judgement whether a user shows a right facial expression matching each displayed emotion.
As we have mentioned in Section 1, there are several datasets that we can use to train the emotion detector. Since our goal is to verify the emotions of users in real scenes, wherever the users are, a natural choice is to use a dataset that is collected in real-world scenes rather than in lab environments. For this reason, we use the CIFE dataset to train the game emotions detector. The CIFE is collected by searching though the Internet and the faces collected are randomly posed, which will make the emotion classifier a robust emotion detector for the seven classes listed in Table \ref{tb1}. We use a CNN-based deep learning approach to construct the emotion classifier, as it has been proved to be effective in several image classification applications ~\cite{xie2015image, cao2015multi, bai2015automatic}. 


The training phase of a deep learning mode using a CNN structure is very time-consuming, since the model has a large number of parameters and need a large dataset to successfully tune those parameters. To save training time, we have decided to use a CNN model that has been already pretrained on a similar dataset, then fine-tune it by training it on additional data ~\cite{li2015emotiw,girshick2014rich}.\textcolor{black}{ We thus chose the Alex model ~\cite{krizhevsky2012imagenet,jia2014caffe} with  8 fully connected (fc) layers to be the pretrained base of our CNN model. The Alex model is a participant in the ImageNet challenge. The number of kernels in layers fc7 and fc8 were 4096 and 1000 in the pretrained model, which was designed for object recognition of 1000 classes. For our application, there are only 7 basic emotion categories, hence the CNN model does not require that many kernels to represent the facial images. Therefore in our CNN structure, we change the number of kernels in fc7 to 2048 and for fc8 to 7 in order to match the number of classes. The parameters of the layers before these two layers are also to be updated.}

After modifying the structure, we  train the new CNN model and fine-tune its parameters using the CIFE dataset. This model can then be used to classify the images captured from  users when playing the game.

\section{Game Design: Interface and Data Collection}

\begin{figure}
\centering
\psfig{file=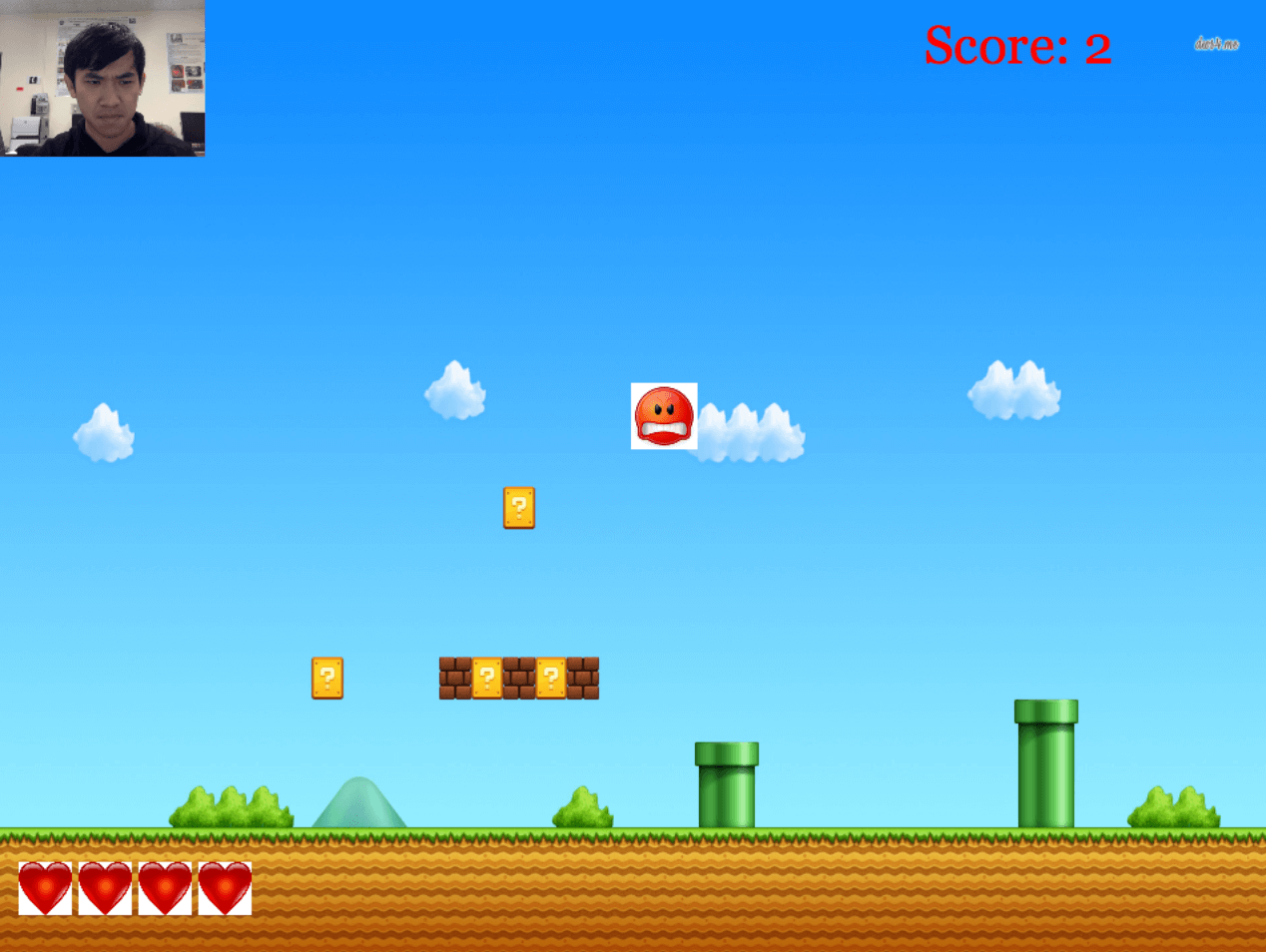, height=2.5in, width=3in,}
\caption{The emotion game scene}
\vskip -6pt
\label{fg2}
\end{figure}

Since we would like game users to show real facial expressions and remain engaged throughout the game, we had to design the emotion game in a way that it is straightforward and interesting as much as possible. After performing a research of the popular web games, we found that the Tower Defense games is the style that fits our task the best. The logic of the Tower Defense games is always very simple: the player needs to build a defense system against the intruders. In our application, we would like the user to act as a defender against an "emotion target". An example of the basic game scene is shown in Figure \ref{fg2}. The live video of the user is shown on the top-left corner of the screen so he can always see his emotion. Each emotion target (a facial emotion icon) will enter the screen as a bomb dropped from the above and the user has to protect the village by making the bomb disappear before it reaches the ground. Some sound effects are also added to make user more engaging. The bomb would disappear if the user makes a facial expression that correctly matches to the displayed emotion (the bomb), as judged by  the CNN-based emotion detector, and the score as shown on the top-right of the screen will be increased.

\begin{figure}
\centering
\psfig{file=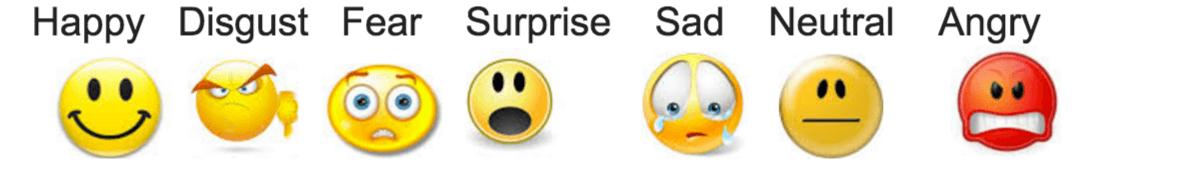, height=0.5in, width=2.5in,}
\caption{Emotion icons used in the game}
\vskip -6pt
\label{fg3}
\end{figure}

However, if the bomb hits the ground, the user will lose one game life, a number of them are shown in the left bottom area of the game scene as red hearts.  A game will be over when there are no more hearts left. As mentioned earlier, each bomb is displayed as a different emotion icon. The seven emotion icons used in this game are illustrated in Figure \ref{fg3}. In order to destroy each bombs, the user's facial expression needs to match the icon on the current screen. 

\subsection {General game design}
Here we would like to provide some technical details of our general game design. The emotion game web interface will access the camera on the user's machine and display the video on the top-left corner of the screen. Then the game interface will capture images of the user's face,  and then send face images to our server. The CNN model we trained in Section 2 will analyze each image and generate a probability vector for the seven emotions and sent it back to the game webpage. The reason for processing face images at a server is due to the high computational  requirements  by the CNN model. After the probability vector of the seven emotions is sent back to the game interface, it compares this feedback with the emotion target ID that has been displayed on the screen as a bomb and informs the server to save the image if it matches the icon emotion. Since the target facial expression that the user needs to make is defined by our system, we not only know the label and have high confidence that facial expression's label is correct, but also can make the dataset more balanced based on our needs. \textcolor{black}{ Our game can be accessed via this test page \footnote { http://emo.vistawearables.com/modeltests/new (Firefox tested and recommended)}.}

The frame rates of typical webcameras are usually 20-60 Hz. We do not need such high frame rates of image capturing for two main reasons: 1) From the computational resource's point of view, the server will have a huge workload if we run the emotion recognition on every single image since the CNN computations are time-consuming.  2) There is no need to know the user's emotion frame by frame. Giving the user some time delay to prepare their emotion may actually result in better image quality and as a consequence, a much better dataset. For these two reasons, we design the game in a way that it only sends one image per second to the server. It takes only around 200 ms for the server to generate the results for every single image, which makes the game run very smoothly. We also set the number of initial game lives to be five and generate the emotion targets randomly, with equal probabilities to all the seven emotions, which theoretically result in a balanced dataset. 

The game is implemented using Javascript and html. The backend is hosted using Ruby on Rails. During the game, emotion icons will drop from the top of the screen, and the user's face is also shown in the left top corner of the screen for the user to check her/his facial emotion. If the user is able to match the emotion before the icon hits the ground, she/he will receive 1 point. The score will change as the user gains or loses points. When all the game lives are used, a "Game Over" sign will be shown up, together with the total score gained by the user,  and then a "Replay" button.

\subsection {Customized game design}
After making the game available to a small group and collecting data from several users who tried it, we realized that the collected dataset is not ideal, specifically for two main reasons: 1) Sometimes it is hard for users to correctly imitate the exact emotions by just looking to the icons; 2) Our emotion detector sometimes is not able to correctly determine whether the subject is making the right face or not. This makes it hard for the players to achieve high scores and as a result, the collected data becomes imbalanced.

We were able to provide two solutions to solve the above problems. 
One is to change the emotion recognition to an emotion verification task. This makes the classification task much easier. Since we know the "ground truth" or the target label for an icon being displayed in the game, we only need to check if the probability of this specific emotion reached a predefined threshold. Each emotion needs to have its own threshold since some emotions are harder to mimic through facial expressions and have higher variety among different users. This will help the users achieve higher scores and also include a broad range of correctly labeled facial expressions for each emotion in the dataset. 

Another solution is to create an individual model for each player based on the CNN extracted features. The user will be compared with her/his individual emotion templates instead of the general CNN model. The Deepface work \cite{taigman2014deepface}  has proved that the CNNs is not only able to directly perform image classification, but can also extract robust features from the images. Thus we extract the features from the CNN for each individual user and then these features are saved as  templates for that specific user. This makes the game customized for each player and the user can gain higher scores and is encouraged to play more.

\begin{figure}
\centering
\psfig{file=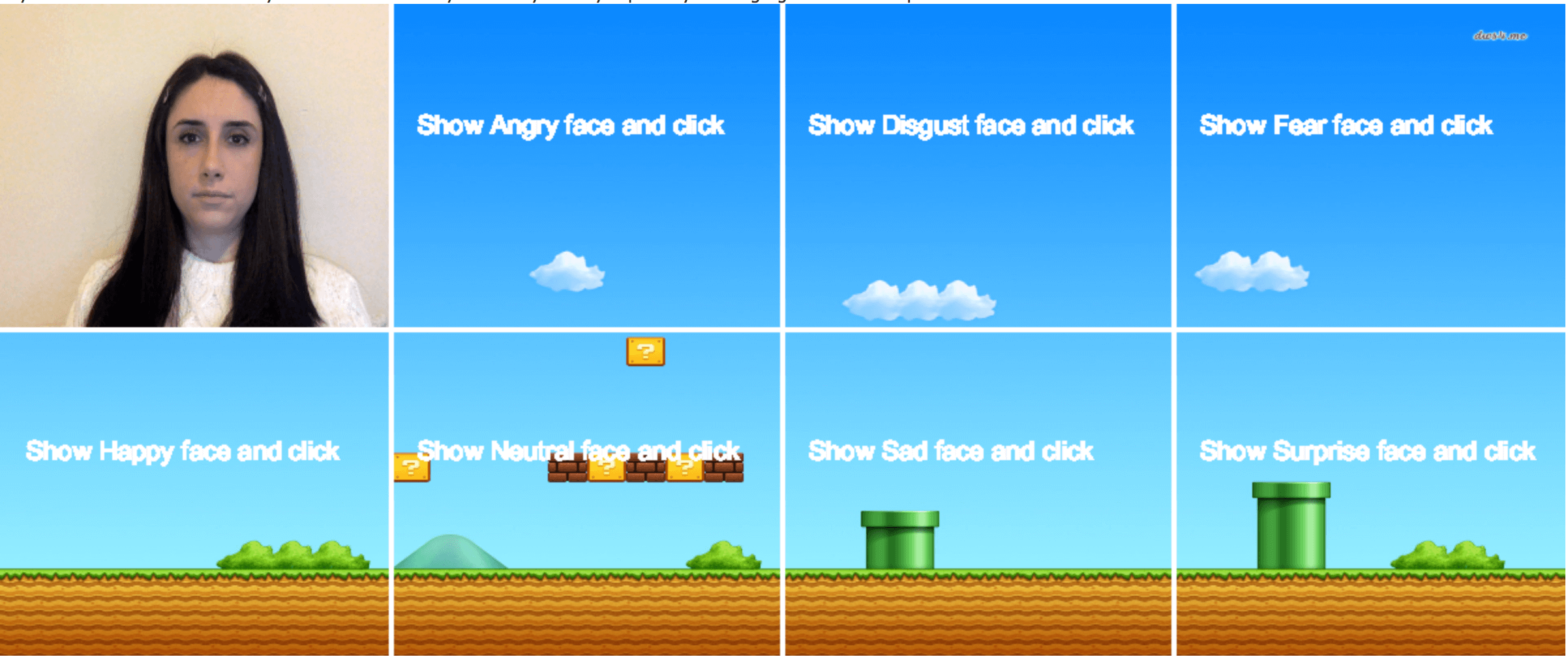, height=1.6in, width=3.2in,}
\caption{A registeration page for the customized game design}
\vskip -6pt
\label{fg4}
\end{figure}

\begin{figure}
\centering
\psfig{file=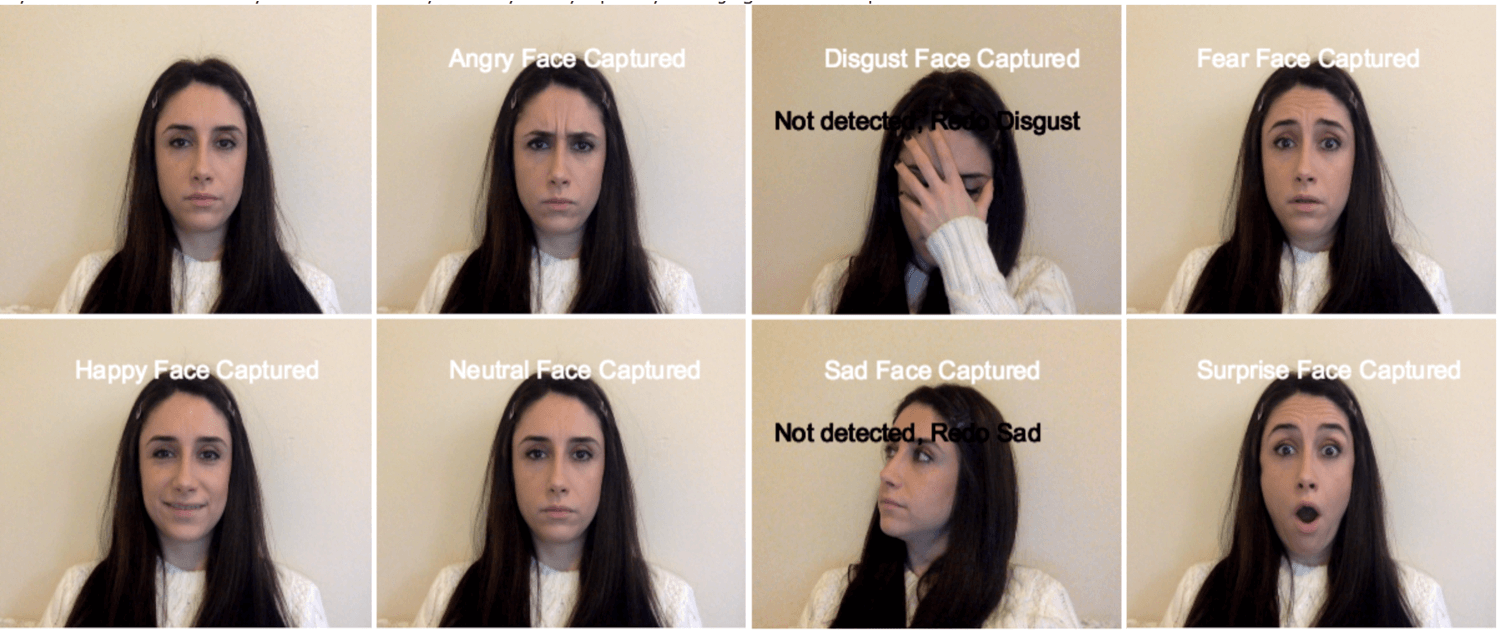, height=1.6in, width=3.2in,}
\caption{Face registration for building individual templates}
\vskip -6pt
\label{fg5}
\end{figure}

As a part of the solutions proposed above, we designed a user registration page as shown in Figure \ref{fg4}. The registration page is divided into 8 subareas. The first subarea shows the current video stream. The other seven subareas display the seven registered emotion templates. To save each template image, the user can click on the corresponding subarea while imitating the correct facial expression. This process can be repeated several times until the user is happy with the saved image. Once all seven emotions are registered, the user can click the "Send All" button to send the emotion templates to the server, where the system will detect the face area in the images and use the CNN model to extract emotion features for the user. If the face cannot be detected, an error message will be sent back to the user, and she/he is then asked to recapture the image for the specific emotion that has caused the error, as shown in Figure \ref{fg5}. \textcolor{black}{ The register page can be accessed here \footnote {http://emo.vistawearables.com/usergames/new (Firefox tested and recommended)}.}

When all the template features are saved in the server, the user will be directed to the customized game scene. While the game is being played, the server will extract features for the image that is being sent at the moment and compare these features to the saved emotion templates. We use L2 distance to select the nearest result and send it back as the detected emotion. Since the features are robust and the user is always compared with her/his own model, the user will potentially achieve a higher score. We call this version of the game the "customized version", as opposes to the previous "general version".


\subsection {Data collection}
\textcolor{black}{ Within one month of the release of the two game test versions to the college students of our department, more than a hundred users played the general version and 74 users tried the customized version.} All the users that we collected data from have signed the consent form of our IRB approval. We obtained more than 15, 000 images in total during this time period and generated the GaMo (game based emotion) Dataset. \textcolor{black}{ Comparing to some deep learning datasets, the size is still not big enough, but our game can run at any time, so we can obtain much bigger dataset when the game reaches more people. The dataset will be made public for research use after the paper is published}. 

 \textcolor{black}{One concern for our dataset might be the use of a trained model to get more emotion data: Will this recognition/ verification model just take emotion data that are similar to our existing data samples and make the dataset less diverse? We believe that by collecting more data from more people, the dateset will be much more diverse. While the general model can contribute to the overall data diversity, the customized version may only collect users specific data similar to their templates. Although  for each individual, a type of emotion tends to be very similar, by assembling all the people together, the data is still diverse. And the most important thing is, deep learning can learn the common features of each emotion well if we can feed all possible data to it. }

We would like to note here that no manual cleanups for the images and labels have been done; all the images are used in our evaluation in Section 4. \textcolor{black}{ By randomly checking the dataset, we have not found any labels that are very off the true expressions. The distribution of the dataset is shown in Table \ref{tb1}. Compared to the CIFE dataset, GaMo is more balanced, which hopefully will result in a much more reliable emotion detector. In conclusion, the data collection is automatic, of high quality and more balanced.  If more images are collected, the numbers of images across the seven categories will be even more balanced.} We will evaluate our dataset in the next section.

\begin{table}
\centering
\caption{Comparison of emotion samples numbers in CIFE and GaMo}

\begin{tabular}{|c|c|c|} \hline
Dataset& CIFE & GaMo \\ \hline
Angry& 1905& 1945\\
Disgust& 975& 1838\\
Fear& 1381& 1586\\
Happy& 3636& 3185\\
Neutral& 2381& 2741\\
Sad& 2485& 1898\\
Surprise& 1993& 2262\\

\hline\end{tabular}
\label{tb1}
\end{table}

\section{Dataset Evaluation}

To determine the usefulness of the GaMo dataset, we performed two experiments. First, using GaMo, we trained a new CNN model by fine-tuning the previous model that has been used in our game engine, which was trained on CIFE . To compare GaMo with CIFE, we ran both a self evaluation and a cross evaluation with the two CNN models: the GaMo CNN model and the CIFE CNN model. Secondly, we designed a small user study to find out if the dataset can actually improve the game engine and game experience. For this purpose, the users played the general version of the game hosted by the two CNN models.
\begin{figure}
\centering
\psfig{file=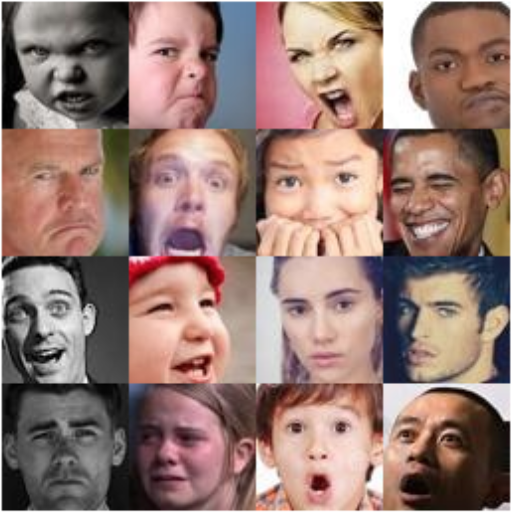, height=2in, width=2in,}
\caption{A few samples from CIFE}
\vskip -6pt
\label{fg6}
\end{figure}
\subsection{CNN model cross validation}
Our goal is to show that the collected balanced dataset can help the classifier achieve better emotion recognition accuracy. To compare the two models, we test the overall accuracy in recognizing all the seven emotions (the Average accuracy) as well as the accuracy of each individual emotion within its own sub-dataset (Angry, Disgust, Fear, Happy, Neutral, Sad, and Surprise). This would give us a good sense on the usefulness of the balanced GaMo dataset. Furthermore, to  compare the performance of the two CNN models to complete new data, we perform a cross evaluation: the model trained on CIFE is tested on images from GaMo and vice versa. 
The results of these experiments are listed in Table \ref{tb2}.

\begin{table}
\centering
\caption{Self and cross evaluation of CIFE and GaMo models}
\begin{tabular}{|c|c|c|c|c|} \hline
     &CIFE& GaMo& CIFE cross& GaMo cross\\ \hline
Average& 0.74&  0.64&  0.21&  0.50\\
Angry&  0.68&  0.65&  0.03&  0.35\\
Disgust&  0.29&  0.57&  0.02&  0.14\\
Fear&  0.44&  0.52&  0.10&  0.36\\    
Happy&  0.87&  0.71&  0.03&  0.80\\
Neutral&  0.75&  0.71&  0.60&  0.33\\
Sad&  0.79&  0.64&  0.17&  0.52\\
Surprise&  0.73&  0.65&  0.09&  0.50\\
\hline\end{tabular}
\label{tb2}
\end{table}

\begin{figure*}
\centering
\psfig{file=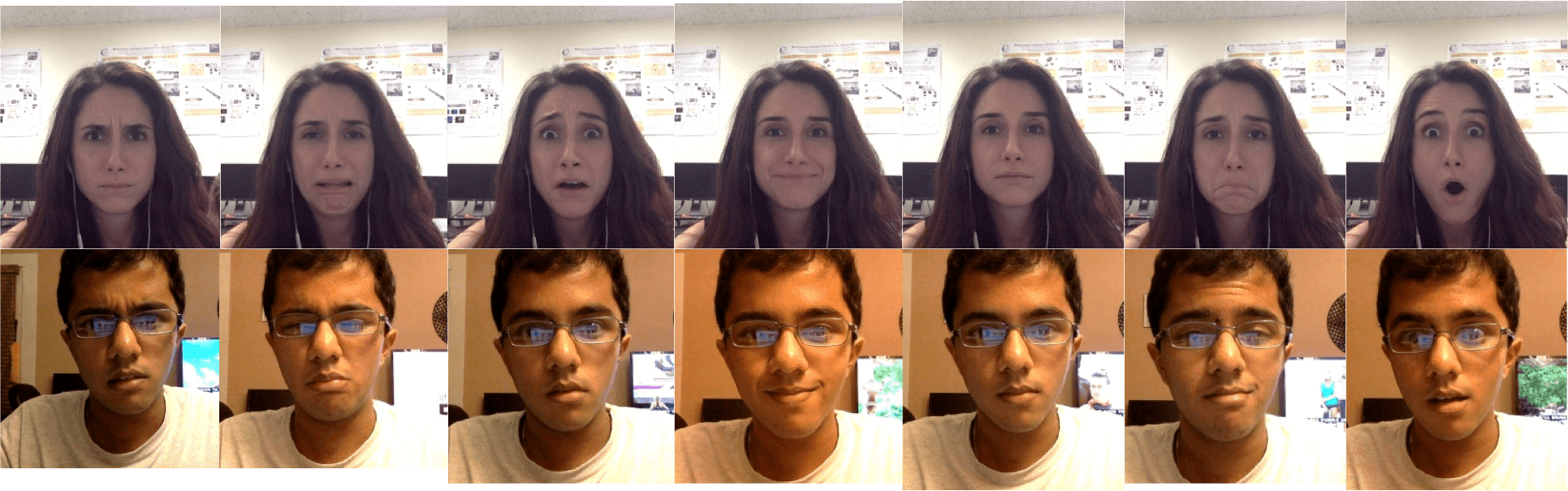, height=2in, width=5in,}
\caption{Comparison of individual template images of two users from GaMo}
\vskip -6pt
\label{fg7}
\end{figure*}

Looking at the self evaluation results, we can see that the model trained on GaMo has a much more balanced distribution on emotion classification on all the seven emotions. Even though the average performance of the CNN model on CIFE is much higher than that on GaMo, the numbers are misleading since the higher average accuracy of the CIFE model is due to the much larger numbers of samples in both Happy and Sad classes, which apparently also have much higher accuracy than others. In comparison, the performance in recognizing Disgust and Fear is much higher using GaMo.

The results of the cross dataset tests are even more interesting. The model trained on CIFE has a very low performance when tested on the GaMo dataset.  We have observed that the difference between the images is significant among the two datasets. Our observations indicate that the emotions in the CIFE dataset tend to be more exaggerated and thus easier to be identified, as it are shown in Figure \ref{fg6}, while the GaMo dataset is more realistic to real life, as it is obtained from ordinary users with a high amount of varieties in imitating facial expressions while playing the game. As an example, Figure \ref{fg7} shows two users who played the game. The first player shows more explicit emotions while the second player's emotions tend to be more implicit. This makes it hard for the model trained on CIFE to classify the images from GaMo. The CIFE model almost completely fails in recognizing Angry, Disgust and Happy in GaMo. We believe the reason is that these three emotions in the CIFE dataset, whether they have fewer or more samples, are much more highly exaggerated than those in the GaMo dataset. On the other hand, when the model trained on GaMo is cross-tested on CIFE, the performance is surprisingly good, even though the performance cannot beat that on the self-test. The reason is that the model is further fine-tuned on a larger, more inclusive and more balanced dataset. The GaMo model does reasonably well on all the three expressions failed by the CIFE model.  In addition, if subtle expressions (as in the GaMo dataset) can be recognized, the exaggerated ones (as in CIFE) are not difficult to detect. As an example, the Happy faces in CIFE can be much more easily recognized (with a 80\% accuracy) using the GaMo model. 

\subsection{Game engine performance study}
\begin{figure}
\centering
\psfig{file=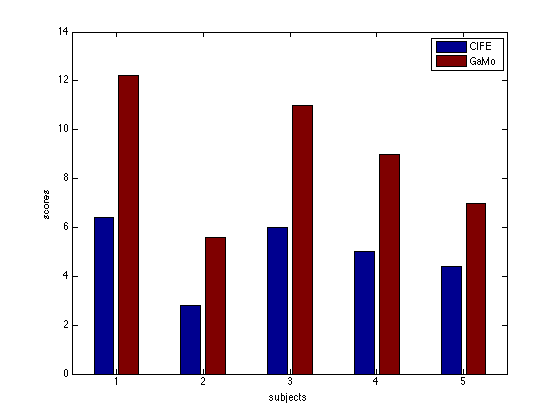, height=2.4in, width=3in,}
\caption{Users average scores on two GaMo and CIFE based CNN models}
\vskip -6pt
\label{fg8}
\end{figure}

The goal of emotion recognition research is often to train a model that can perform well in real scenes. This is especially true in human-computer interaction and multimedia retrieval applications of real daily activities, such as satisfaction studies of customers and viewers, and assistive social interaction for people in need, such as individuals with visual impairment and Autism spectrum disorders. One approach to verify an emotion detector is through a test on ordinary people with natural facial expressions. To accurately evaluate the two models, we analyze the data collected from five new users \textcolor{black}{ (3 male and 2 female)} who are not included in GaMo, while playing the general version of the game. Note the in the phase of GaMo data collection, we mainly use the customized game interface since users cannot perform well with the general game interface. In this game engine performance study, the general game is played five times by each user \textcolor{black}{with the same game settings} and the scores  of the five rounds are recorded, using the two versions of our game engine, one trained on CIFE and the other on GaMo, respectively. Figure \ref{fg8} shows the result of this experiment. We have plotted the two average scores for each player on games powered by the two game engines. According to this figure, the GaMo game engine has a much better performance and results in higher scores. This further confirms that the model trained on GaMo is more suitable for real-world emotion recognition.

\begin{figure}
\centering
\psfig{file=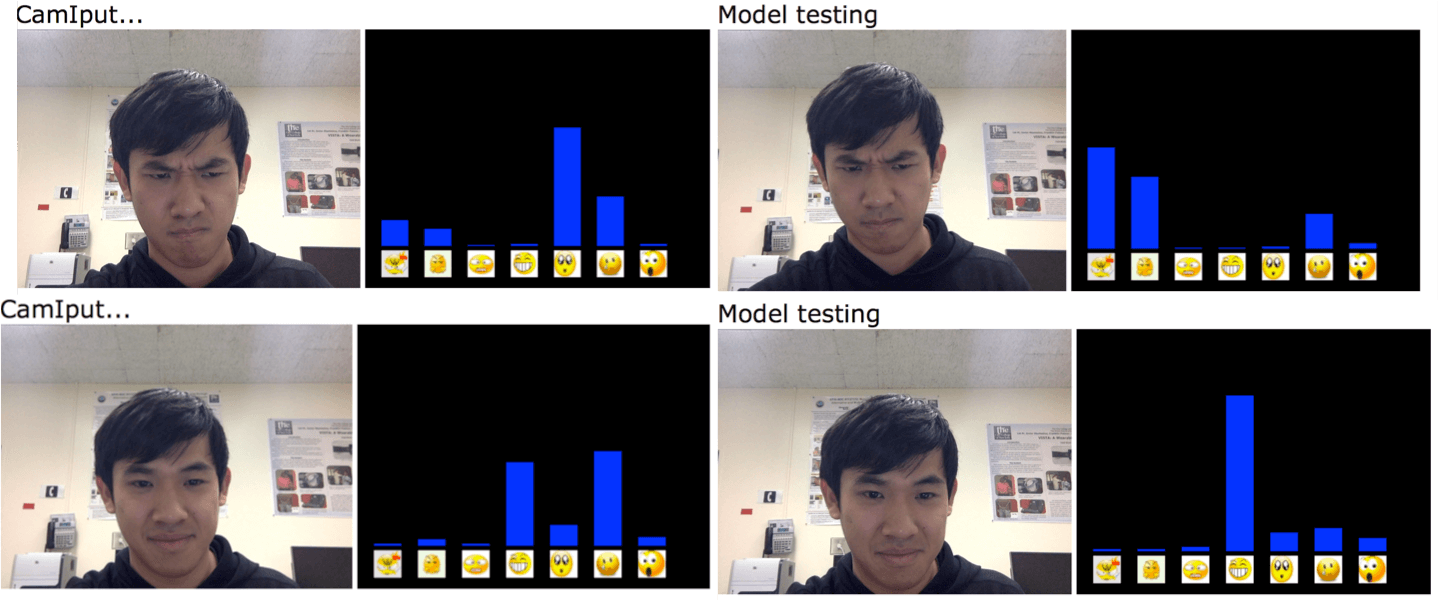, height=2in, width=3.2in,}
\caption{User in game shots (left two are from the CIFE model and right two are from the GaMo model). Bars show probability of each emotions. The order of the emotions is Angry, Disgust, Fear, Happy, Neutral and Surprise. For some subtle emotions, only the GaMo model works well.}
\vskip -6pt
\label{fg9}
\end{figure}

This  result agrees with the cross testing results which show that the GaMo model has a better performance on the GaMo dataset itself. These observations would also support our claim that GaMo is very useful in detecting subtle emotions. For instance, the user can gain a point with a normal smile expression in GaMo model game as shown in Figure \ref{fg9}, while in the CIFE model, the expression can not be detected. Same fact holds for detecting anger or any other emotions, as our players do not have any prior knowledge of how obvious and explicit their facial expression should look like. 

\section{Conclusion}
In this paper, we have presented a multimedia gamification based framework for balanced emotion dataset generation. Based on the framework, we developed two versions of the game for the users to play: a general one (which uses facial emotion recognition and is more challenging) and a customized one (which uses facial emotion verification and is less challenging). Both interfaces have been used for collecting our new emotion dataset, which we have named GaMo. The GaMo dataset is automatically built while users are playing any of these two versions of the game. We successfully collected more than 15,000 images and trained a fine-tuned CNN model on the GaMo dataset. The evaluation results show that the CNN model trained on GaMO achieves higher and more balanced performance compared to the CNN model trained on CIFE, which is a web-collected emotion dataset. The model trained on GaMo is able to detect subtle emotions with much higher accuracy that a similar model trained on CIFE.  Thanks to the web-based gamification interface for facial emotion data collection, we will be able to collect more and higher quality data much more efficiently. Therefore in the future work, we will investigate the impact of more facial emotion data on the performance of facial expression recognition. We would also like to apply the interface to some real-world applications, such as social interaction for visually impaired people and people with Autism Spectrum Disorders (ASDs).
\section{Acknowledgement}
This work is supported by the National Science Foundation through Award EFRI -1137172, and VentureWell (formerly NCIIA) through Award 10087-12.
{\small
\bibliographystyle{ieee}
\bibliography{egbib}
}

\end{document}